\documentclass[10pt,twocolumn,letterpaper]{article}

\usepackage{cvpr}
\usepackage{times}
\usepackage{epsfig}
\usepackage{graphicx}
\usepackage{amsmath}
\usepackage{amssymb}
\usepackage[utf8]{inputenc} 
\usepackage[T1]{fontenc}    
\usepackage{url}            
\usepackage{booktabs}       
\usepackage{amsfonts}       
\usepackage{nicefrac}       
\usepackage{microtype}      
\usepackage{amsmath,amssymb,bbm}
\usepackage{graphicx, color}
\usepackage{subfigure}
\usepackage{caption}
\usepackage{xspace}
\usepackage{algorithm}
\usepackage[noend]{algcompatible}
\usepackage[bottom]{footmisc}
\usepackage{tablefootnote}

\usepackage[pagebackref=true,breaklinks=true,letterpaper=true,colorlinks,bookmarks=false]{hyperref}

\cvprfinalcopy 

\ifcvprfinal\fi

\graphicspath{{./Figures/}}

\begin{document}
\title{Divide, Denoise, and Defend against Adversarial Attacks}
\author{
  Seyed-Mohsen Moosavi-Dezfooli$^{1,2}$ \\
  \tt\small{seyedmoosavi@apple.com} \\
  \and
  Ashish Shrivastava$^1$ \\
  \tt\small{ashish.s@apple.com} \\
  \and
  Oncel Tuzel$^1$ \\
  \tt\small{otuzel@apple.com} \\
  $^1${Apple Inc., USA.}   \;\;\;  $^2${\'Ecole Polytechnique F\'ed\'erale de Lausanne, Switzerland.} \\
}

\maketitle

\begin{abstract}
  Deep neural networks, although shown to be a successful class of machine learning algorithms, are known to be extremely unstable to adversarial perturbations.
  Improving the robustness of neural networks against these attacks is important, especially for security-critical applications.
  To defend against such attacks, we propose dividing the input image into multiple patches, denoising each patch independently, and reconstructing the image, without losing significant image content.
  We call our method D3.
  This proposed defense mechanism is non-differentiable which makes it non-trivial for an adversary to apply gradient-based attacks.
  Moreover, we do not fine-tune the network with adversarial examples, making it more robust against unknown attacks.
  We present an analysis of the tradeoff between accuracy and robustness against adversarial attacks.
  We evaluate our method under black-box, grey-box, and white-box settings.
  On the ImageNet dataset, our method outperforms the state-of-the-art by $19.7\%$ under grey-box setting, and performs comparably under black-box setting.
  For the white-box setting, the proposed method achieves $34.4\%$ accuracy compared to the $0\%$ reported in the recent works.
\end{abstract}

\section{Introduction}
Deep neural networks (DNNs) have produced valuable results on many practical applications~\cite{mscoco, imagenet_cvpr09, ZhangLL16, youtub8M_16,openimages},
but are vulnerable to even small adversarial perturbations~\cite{goodfellow15,NarodytskaK16,Brown17}.
In particular, such perturbations can change the decision of DNN-based image classifiers.
The vulnerability of deep networks to adversarial manipulations of their input goes beyond classification tasks and additive perturbations \cite{cisse17, metzen17_sem_seg,xiao2018spatially,Kanbak2017}.
Moreover, the attacks are transferrable, meaning that an adversary can find these perturbations without having access to the network.
For example, \cite{song16,brendel2018decisionbased} successfully attacked image classifiers used in commercial applications.
These observations highlight the need to improve the robustness of deep networks, especially, if they are deployed in a hostile or security-critical environment.

Many existing defense methods either show results on small datasets, such as MNIST or CIFAR-10~\cite{madry17,mittal17,samangouei2018defensegan}, or they add adversarial examples to the training data~\cite{tramer17}.
Since new attacks are constantly being proposed, the defense should be attack-agnostic to make it robust against an unknown attack.
An ideal defense should be non-differentiable so it does not allow the adversary to back-propagate through the defense mechanism.

\begin{figure}[t]
  \includegraphics[width=0.9\columnwidth]{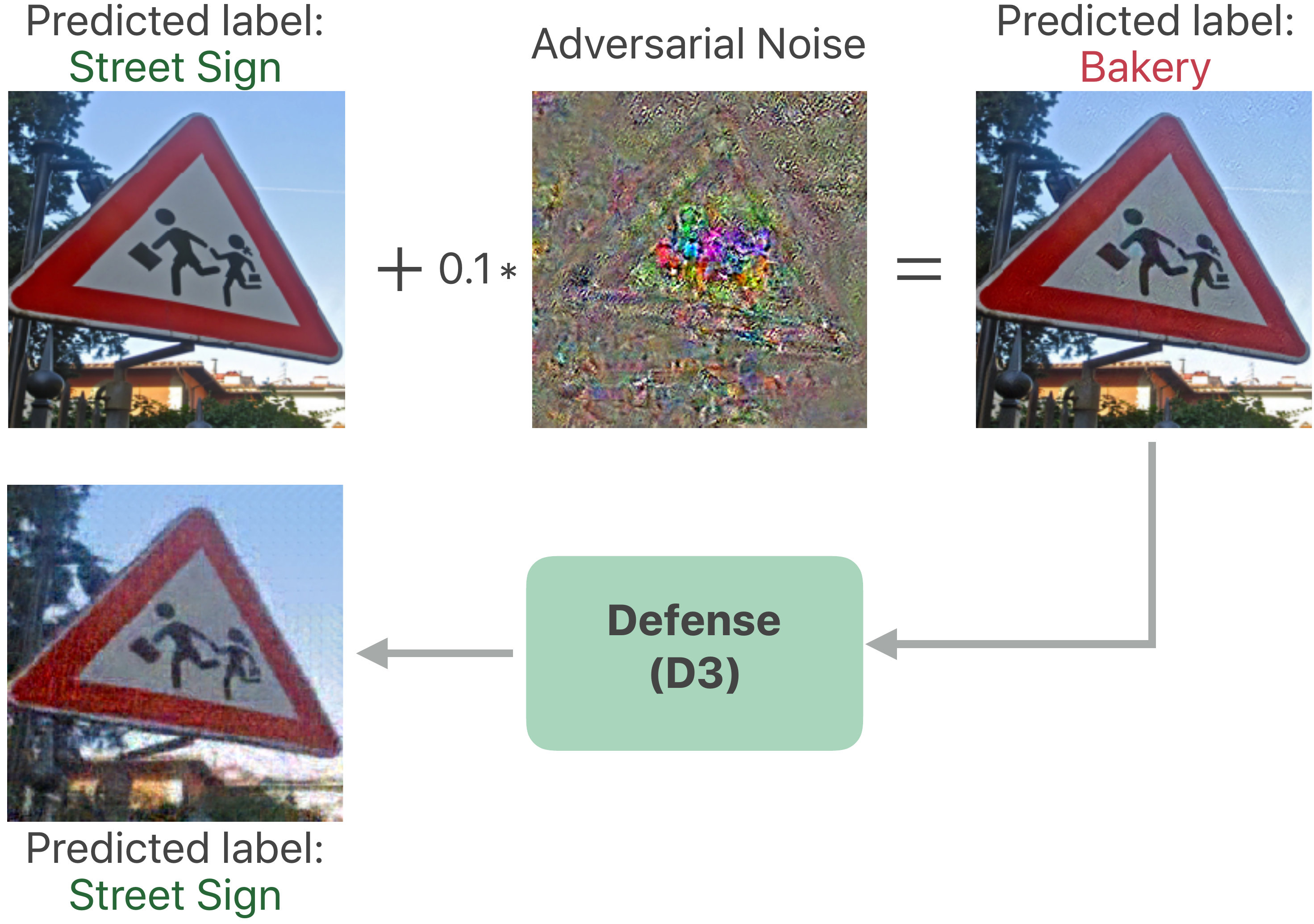}
  \vskip-0.1in
  \caption{The proposed method (D3) transforms the input image using a non-differentiable algorithm.
  This transformation removes adversarial noise to improve robustness to attacks. }
  \label{fig:defense}
\end{figure}

Our defense mechanism (Figure~\ref{fig:defense}) maps an input image to a new space and is based on the following observations:
(1)~increased dimensionality has an adverse effect on the robustness of deep networks~\cite{fawzi16},
(2)~the mapping should not reduce the accuracy of clean data, and
(3)~the mapping should be stable such that the output is minimally sensitive to input perturbations.

Our method divides the input image into multiple overlapping patches that are projected to a lower dimensional space using a dictionary.
The dictionary consists of clean image patches to reconstruct the input image patches with a variant of the matching pursuit (MP) algorithm~\cite{Mallat_1993}.
We use this algorithm for denoising because it is fast and non-differentiable.
We propose a novel patch-selection algorithm to construct the dictionary such that the selected patches are not too similar and they represent the salient parts of the image.
This selection process mitigates the effect of adversarial perturbations while maintaining good accuracy on clean images.

We evaluate our algorithm using the ImageNet dataset under three settings --
(1)~black-box attacks where the adversary does not know about the network or the defense method,
(2)~grey-box attacks where the adversary knows the network parameters but not about the defense mechanism, and
(3)~white-box attacks where the adversary knows both the network parameters and the defense algorithm.
Our algorithm performs comparably to other state-of-the-art methods on the black-box setting and performs significantly better on grey-box and white-box attacks.
In addition, we show that as task complexity decreases (\eg with a subset of the ImageNet classes), we can remove more information content from the image by denoising while maintaining high accuracy and robustness.
Our contributions are:
\vspace{-0.1in}
\begin{itemize}
\setlength\itemsep{0.1em}
\item We propose a novel framework for defending against adversarial attacks by dividing images into overlapping patches and denoising them independently using a non-differentiable, attack-agnostic algorithm. Our patch-based method is particularly designed for high-dimensional datasets, e.g. ImageNet.
\item We provide a thorough analysis of the tradeoff between clean image accuracy and robustness against adversarial attacks.
\item We extensively evaluate our method on the ImageNet and CIFAR-10 datasets under different adversarial settings.
We compare with the state-of-the-art papers and the NIPS-2017 challenge, and show improved performance.
\end{itemize}


\section{Related Work}
The seminal work by~\cite{szegedy14} highlights the vulnerability of deep neural networks to adversarial examples.
Since then, many methods have been proposed to assess such vulnerability by developing various adversarial attacks.
In~\cite{goodfellow15}, the authors proposed Fast Gradient Sign Method (FGSM) which attacks a classifier by computing the sign of the gradient of the loss \wrt the input images.
To assess the robustness of deep networks more accurately, iterative algorithms such as DeepFool~\cite{moosavi16} and C\&W ~\cite{carlini17} have later been introduced.
It is also possible to use generative models like Generative Adversarial Networks (GANs) ~\cite{Goodfellow14} to generate adversarial perturbations ~\cite{fischer17_1,Hayes2017}.

In~\cite{szegedy14}, it has also been shown that adversarial attacks are transferrable and can be used in black-box settings.
In these settings, the adversary has access neither to the weights of the network nor to the architecture.
More recently, it has been shown that there exists image-agnostic attacks, universal adversarial perturbation~\cite{moosavi17}, which can be added to any image to fool a given network.
Even worse, these perturbations can be computed without the dataset used for training the network~\cite{mopuri2017fast}.

In recent years, there have been several efforts to defend against adversarial attacks.
The first defense against adversarial perturbations was proposed by~\cite{gu14} where they use stacked denoising auto-encoders to mitigate perturbations. A similar approach has been studied in \cite{meng2017} to denoise adversarial examples. Recently, other generative models, e.g. GANs and PixelCNN~\cite{vandenoord2016}, have been used to project back the malicious samples on the manifold of data~\cite{samangouei2018defensegan,song2018pixeldefend}. However, such methods are restricted to small-scale datasets such as MNIST and CIFAR-10.
In~\cite{papernot16}, distillation was suggested as a defense; however, it only masks the gradient and is still vulnerable in black-box settings as demonstrated in~\cite{carlini17}.
In \cite{madry17}, authors provide an efficient adversarial training framework based on the robust optimisation to counter the first-order adversaries. However, their method is not model-agnostic and due to computational complexities, it has not been applied on large-scale datasets such as ImageNet. Recently, in~\cite{tramer17}, adversarial training has been applied on ImageNet using an ensemble of networks.
The main drawback of such adversarial training scheme is that it overfits the noise and does not generalize well against an unknown attack~\cite{guo17}.
Very recently, several strategies to defend against attacks have been explored using various image transformations~\cite{guo17}.
As a different approach, detecting malicious samples, instead of improving the robustness, is sought in~\cite{metzen17,Feinman2017}.
They demonstrated that deep networks can be augmented with a network to detect adversarial examples.
This approach also suffers from overfitting to  specific types of perturbations.
Recent work,~\cite{carlini17_2}, has successfully attacked ten different defense strategies emphasizing the difficulty of this problem.

There have been few theoretical works studying the robustness of deep networks.
There is a tradeoff between accuracy and robustness of kernel classifiers~\cite{fawzi15}.
In~\cite{fawzi16}, the authors established a bound on the robustness of a certain type of classifiers when the adversary is restricted to a low dimensional space.
In~\cite{sinha17, HeinA17}, some lower bounds have been derived on the robustness of simple neural networks.

\begin{figure}[t]
  \includegraphics[width=0.9\columnwidth]{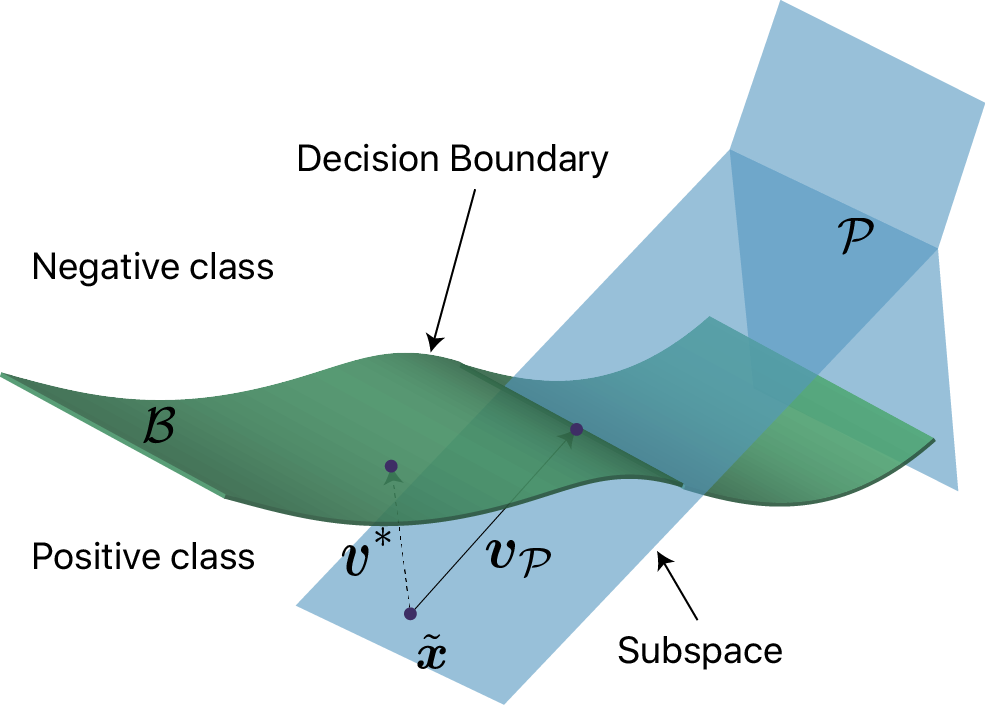}
  \vskip-0.1in
  \caption{Reducing dimensionality improves the robustness of the classifier.
  $\mathcal P$ is a union of subspaces illustrated by the blue hyper-planes.
  Here, $\tilde{\boldsymbol x} = T(\boldsymbol x)$ is the image projected to the nearest subspace in  $\mathcal P$.
  To avoid clutter, we are not showing the original image $\boldsymbol x$.
  The adversarial noise $\boldsymbol v^*$ is the smallest distance from $\tilde {\boldsymbol x}$ to the classifier's decision surface $\mathcal B$.
  When the adversary is restricted to a smaller dimensional subspace (the projected hyper-plane), the norm of the noise $\boldsymbol v_{\mathcal P}$ is much bigger than the norm of $\boldsymbol v^*$ to cross the decision boundary.
  }
  \label{fig:dim_red_justification}
\end{figure}

\begin{figure}[t]
  \centering
  \includegraphics[width=1.0\columnwidth]{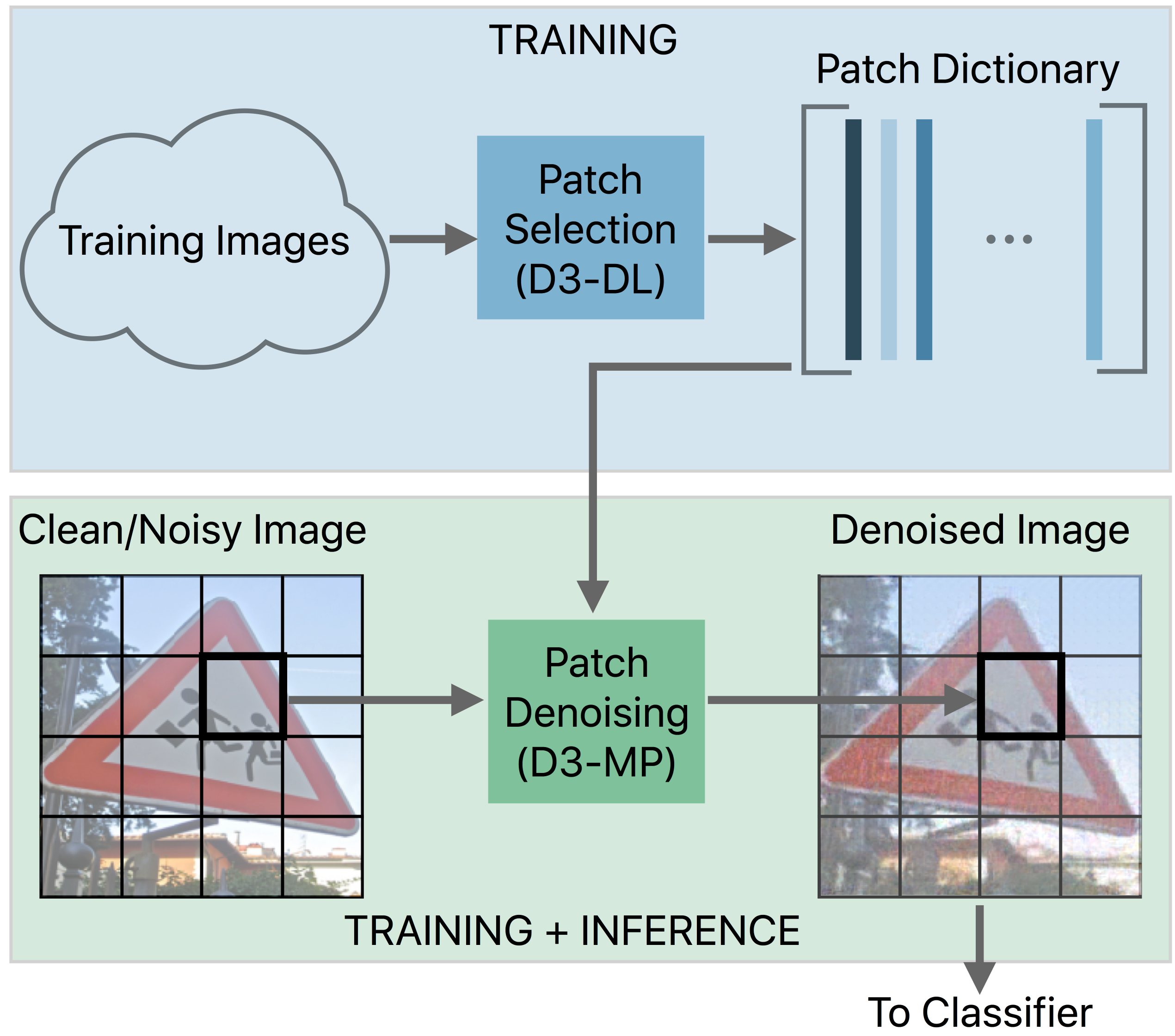}
  \vskip-0.15in
  \caption{Overview of the proposed D3 algorithm.
  An input image is divided into patches and each patch is denoised independently.
  }
  \label{fig:overview}
\end{figure}

\section{Problem Formulation}
Let $f_{\boldsymbol \theta}(\boldsymbol x): \mathbb R^d \rightarrow \mathbb N$ be a classifier parameterized by $\boldsymbol \theta$ that computes the class of an input image $\boldsymbol x$,
where $\mathbb N$ is the set of natural numbers denoting class labels.
An adversary can perturb the image with noise $\boldsymbol v$ such that $f_{\boldsymbol \theta}(\boldsymbol x) \neq f_{\boldsymbol \theta}(\boldsymbol x + \boldsymbol v)$.
The robustness of the classifier at $\boldsymbol x_0$, denoted by $\rho (\boldsymbol x_0)$, can be defined as the minimum perturbation needed to change the predicted label~\cite{fawzi16}:
\begin{align}
 \rho(\boldsymbol x_0) = \arg\min_{\boldsymbol v} \|\boldsymbol v\|_2, \;\text{s. t.}\;\; f_{\boldsymbol \theta}(\boldsymbol x) \neq f_{\boldsymbol \theta}(\boldsymbol x + \boldsymbol v).
\end{align}
The noise can be scaled to make the attack stronger.
We improve robustnesss by learning a stable transformation, $T(.)$, such that the label of the transformed image does not change when corrupted with the noise,
\ie $f_{\boldsymbol \theta}(T(\boldsymbol x)) = f_{\boldsymbol \theta}(T(\boldsymbol x + \boldsymbol v))$.
Moreover, we want to maintain the accuracy when the clean images are transformed by $T(.)$ ensuring that the $f_{\boldsymbol \theta}(T(\boldsymbol x))$ is equal to the ground truth label of $\boldsymbol x$.
Most attacks rely on computing gradients of the classification function \wrt the input.
Hence, it is desirable to make the transformation $T(.)$ non-differentiable so that the gradients cannot pass through the defense block.
To create a defense mechanism that is robust against an unknown future attack, we want to keep the defense algorithm to be attack-agnostic and do not want to fine-tune the network with simulated adversarial images.

Under some regularity conditions, restricting the adversary to a low dimensional subspace can improve robustness~\cite{fawzi16}.
Simple dimensionality reduction methods, such as PCA, have been studied in \cite{mittal17,HendrycksG16b} to improve defense against adversarial perturbations.
However, such solutions only work on simpler tasks (such as classification on MNIST digits) as they significantly decrease the discriminative performance of the network.
Such transformations usually remove the high frequency information required for complex tasks such as $1000$-class classification on ImageNet dataset~\cite{imagenet_cvpr09}.
Therefore, a better information preserving dimensionality reduction method is required to limit the space of adversarial noise, while keeping important details.

\section{The D3 Algorithm}
\label{sec:D3_algorithm}
Assume that an operator $T(\boldsymbol x)$ projects the input image $\boldsymbol x \in \mathbb R^d$ to the closest subspace in a union of $m$ dimensional subspaces.
This operation is a linear projection operator onto an $m$ dimensional subspace in a local neighborhood of $\boldsymbol x$.
For additive perturbations, the adversary is limited to \textit{locally} seeking noise in an $m$ dimensional subspace and the robustness, $\rho$,
can ideally be improved by a factor of $\sqrt{d/m}$~\cite{fawzi16}.
This intuition is illustrated in Figure~\ref{fig:dim_red_justification}.
Motivated by this result, we look for a transformation, $T(.)$, satisfying the following conditions:
(1)~$\text{rank } J_T(\boldsymbol x) \ll d$, where $\boldsymbol x\in\mathbb{R}^d$, where $J_T(\boldsymbol x)$ is the Jacobian matrix in a small neighborhood of $\boldsymbol x$, and
(2)~$\|T(\boldsymbol x)-\boldsymbol x\|$ is small.
The first condition ensures that the local dimensionality is low, and
the second condition means that the image and its transformed version are close enough to each other to preserve visual information.

We propose a patch-based denoising method for defense.
We divide the input image into multiple patches, and denoise them independently with sparse reconstruction using a dictionary of patches.
Assume that $\kappa$ is the sparsity (number of components used to reconstruct a patch) and each patch is $P \times P$ pixels.
Then, for non-overlapping patches, the local dimensionality of our projection operator $T(.)$ would be $\kappa \frac{d}{P^2}$.
According to \cite{fawzi16}, this dimensionality reduction would ideally improve robustness by a factor of $\frac{P}{\sqrt{\kappa}}$.

Sparse reconstruction and dictionary-based methods have been widely used to enhance the quality of images~\cite{Aharon2006_ksvd, Mairal_2008, Elad_2010, Wright_2010, Shrivastava2015, Pati93orthogonalmatching}.
For computational efficiency, we use sampled image patches as our dictionary.
We use a novel patch selection algorithm that is optimized to improve robustness of the classifier.
For sparse reconstruction, we use an efficient greedy algorithm which is a variant of matching pursuit~\cite{Mallat_1993} .
Our method is summarized in Figure~\ref{fig:overview}.



Computing the performance of the D3 algorithm for different hyper-parameters requires both training the D3 algorithm and the classification network, which are computationally expensive.
Therefore, to efficiently study the effect of hyper-parameters of our algorithm, we compute the following proxy metrics:\\
(1) \textbf{``Matching-rate''} (MR) is the fraction of patches that are identical in the denoised image $T(\boldsymbol x + \boldsymbol v)$ and the clean image $T(\boldsymbol x)$.
Let $\{\boldsymbol{p}_1,\boldsymbol{p}_2,\dots,\boldsymbol{p}_n\}$ and $\{\hat{\boldsymbol{p}}_1,\hat{\boldsymbol{p}}_2,\dots,\hat{\boldsymbol{p}}_n\}$ be patches extracted from $\boldsymbol{x}$ and $\boldsymbol{x}+\boldsymbol{v}$, respectively.
The matching-rate is defined as, $\text{MR} = \mathbb{E}_{\boldsymbol{x}\in\mathcal{D}}(\gamma(\boldsymbol{x}))$, where
$$
  \gamma(\boldsymbol{x})=\frac{1}{n}\sum_{j=1}^n\mathbbm{1}_{\left\|T\left(\boldsymbol{p}_j\right)-T\left(\hat{\boldsymbol{p}_j}\right)\right\|_\infty \leq \delta}
  $$
and $\mathbbm{1}_{[.]}$ is an indicator function. Here, with slight abuse of notation, we assume that $T(.)$ is applied to patches.
Higher MR corresponds to being more robust to the attacks. \\
(2)\textbf{``Reconstruction-error''} (RE) is the average $\ell_2$ distance between the clean image, $\boldsymbol x$, and the transformed image, $T(\boldsymbol x)$:
 $\text{RE} = \mathbb{E}_{\boldsymbol{x}\in\mathcal{D}}(\|\boldsymbol{x}-T(\boldsymbol{x})\|_2/\|\boldsymbol{x}\|_2).$
A higher reconstruction-quality, ($1 - \text{RE}$), results in higher classification accuracy for the clean images as more information is retained. 

Our experiments show that these proxy metrics are highly correlated with accuracy and robustness of the classifier.
We use $500$ images randomly chosen from the dataset to quickly compute these values.
Next, we describe the patch denoising algorithm (D3-MP), and the patch-selection algorithm to construct the dictionaries (D3-DL).

\subsection{Patch-based Denoising (D3-MP)}
Let $\{\boldsymbol{S}_i\}_{i=1}^{\kappa}$ be a set of dictionaries computed using our patch-selection algorithm.
Each dictionary $\boldsymbol S_i \in \mathbb R^{P^2 \times \eta}$ is a matrix
containing $\eta$ columns of dimension $P^2$.
Here, $\kappa$ is the sparsity.
The first dictionary $\boldsymbol S_1$ is used to select the first atom, while reconstructing a given patch $\boldsymbol p$.
Then, the residual is computed between the image patch, $\boldsymbol p$, and the selected atom, $\boldsymbol s_l$.
As in the standard matching pursuit (MP), the residual is used to select the next atom.
But unlike standard MP, we use a different dictionary, $\boldsymbol S_i$, to select at the $i^\text{th}$ sparsity level.
We provide pseudo-code for this approach in Algorithm~\ref{alg:D3MP}.

%

%
%
%

\subsection{Patch Selection Algorithm to Learn Dictionaries (D3-DL)}
To scale up the dictionary learning task for a large dataset such as the ImageNet, we propose an efficient greedy patch-selection algorithm. 
As mentioned earlier, we compute multiple dictionaries for different sparsity levels in our D3-MP algorithm.

We build the set of dictionaries in a greedy manner by selecting the ``important'' and ``diverse'' patches.
The algorithm takes into account the saliency information of images.
The norm of the gradient of the classification function \wrt to the input image is used as the saliency map.
We do importance sampling among all the patches \wrt the saliency map.
We add a patch to the dictionary if the reconstruction of the patch using the existing dictionary has greater than a threshold angular distance, $\epsilon$, from the patch.
The saliency map helps to preserve the details that are important for the classification task, and the cutoff on the angular distance ensures that the dictionary is diverse.

In our experiments, using a pre-tained network on the ImageNet dataset, we find $4\%$ improvement in classification accuracy with the saliency map compared to randomly selecting a patch from the whole image.
The diversity among dictionary atoms encourages mapping a clean and corresponding noisy image patch to the same dictionary atom.
Ensuring that any two patches from the dictionary are a certain threshold apart also improves the MR
and the robustness of the classifier.

After the first dictionary is constructed, we reconstruct the image patches using this dictionary and compute the residuals.
The next dictionary is constructed on the residual images instead of the original images.
This process is repeated for all the remaining dictionaries (see Algorithm~\ref{alg:D3PatchSelection} for pseudo-code).
We found that the MR and $1$ - RE were higher when we used a different dictionary for different sparsity levels using residual images compared to using one common dictionary for all sparsity levels constructed from original images.
For example, with $\kappa=2$, we found the MR = $0.88$, $1$ - RE = $0.83$ when using two separate dictionaries where the second dictionary contained residuals instead of image patches.
Comparing those to MR = $0.80$, $1$ - RE = $0.81$ when using one dictionary, constructed using original image patches, shows that using multiple dictionaries gives better matching-rate and reconstruction quality.

\begin{algorithm}[h]
   \caption{D3-MP}
   \label{alg:D3MP}
\begin{algorithmic}
\STATE {\bfseries Input:} A set of dictionaries $\{\boldsymbol{S}_i\}_{i=1}^{\kappa}$, image patch $\boldsymbol{p}$.
\STATE {\bfseries Output:} processed patch $\boldsymbol{q}$.
  \STATE $\boldsymbol{q}\leftarrow \boldsymbol{0}$
  \STATE $\hat{\boldsymbol{p}}\leftarrow \boldsymbol{p}$
  \FOR{$i=1$ \bfseries{to} $\kappa$}
    \STATE $\boldsymbol{a} \leftarrow \hat{\boldsymbol{p}}^\top\boldsymbol{S}_i$
    \STATE $l \leftarrow \text{argmax}_k |a_k|$
    \STATE $\boldsymbol{q} \leftarrow \boldsymbol{q}+a_l\boldsymbol{s}_l$
    \STATE $\hat{\boldsymbol{p}} \leftarrow \hat{\boldsymbol{p}}-a_l\boldsymbol{s}_l$
  \ENDFOR
  \STATE \bfseries{return} $\boldsymbol{q}$
\end{algorithmic}
\end{algorithm}%
\begin{algorithm}[h!]
   \caption{D3-DL}
   \label{alg:D3PatchSelection}
\begin{algorithmic}
\STATE {\bfseries Input:} saliency algorithm $\mathcal{H}$, training images $\mathcal{D}$, size of dictionary $\eta$, sparsity $\kappa$, $\epsilon$.
\STATE {\bfseries Output:} set of dictionaries $\{\boldsymbol{S}_i\}_{i=1}^{\kappa}$.
\FOR {{$i=1$ \bfseries{to} $\kappa$}}
  \STATE $n\leftarrow 0$
  \STATE $\boldsymbol{S}_i\leftarrow []$
  \WHILE {$n < \eta$}
    \STATE Randomly select $\boldsymbol{x}\in\mathcal{D}$.
    \STATE Compute saliency map $\mathcal{H}(\boldsymbol{x})$.
    \STATE Randomly select patch $\boldsymbol{s}$ from $\boldsymbol{x}$ according to $\mathcal{H}(\boldsymbol{x})$.
    \IF{$i=1$}
      \STATE $\tilde{\boldsymbol{s}}\leftarrow$ D3-MP$\left(\{\boldsymbol{S}_1\},\boldsymbol{s}\right)$
      \STATE  //Add $\boldsymbol{s}$ to $\boldsymbol{S}_1$ if it is arcsin($\epsilon$) away.
      \IF{$\|\boldsymbol{s}-\tilde{\boldsymbol{s}}\|_2/\|\boldsymbol{s}\|_2>\epsilon$}
        \STATE Concatenate $\boldsymbol{s}/\|\boldsymbol{s}\|_2$ to columns of $\boldsymbol{S}_i$
        \STATE $n\leftarrow n+1$
      \ENDIF
    \ELSE
      \STATE $\tilde{\boldsymbol{s}}\leftarrow$ D3-MP$\left(\{\boldsymbol{S}_j\}_{j=1}^{i-1},\boldsymbol{s}\right)$
      \STATE $\boldsymbol{r}\leftarrow\boldsymbol{s}-\tilde{\boldsymbol{s}}$
      \STATE  //Add $\boldsymbol{r}$ to $\boldsymbol{S}_i$ if it is arcsin($\epsilon$) away.
      \STATE $\tilde{\boldsymbol{r}}\leftarrow$ D3-MP$\left(\{\boldsymbol{S}_i\},\boldsymbol{r}\right)$
      \IF {$\|\boldsymbol{r}-\tilde{\boldsymbol{r}}\|_2/\|\boldsymbol{r}\|_2>\epsilon$}
        \STATE Concatenate $\boldsymbol{r}/\|\boldsymbol{r}\|_2$ to columns of $\boldsymbol{S}_i$
        \STATE $n\leftarrow n+1$
      \ENDIF
    \ENDIF
  \ENDWHILE
\ENDFOR
\end{algorithmic}
\end{algorithm}

\subsection{Denoising Algorithm}
\label{sec:D3_denoising_algo}
The proposed defense algorithm (D3): (1) divides the input image into overlapping patches, (2) denoises each patch (with D3-MP) using the constructed dictionaries (with D3-DL), and (3) reconstructs the denoised image by averaging the pixels in overlapping patches.
In our experiments, we set the amount of overlap to $75\%$ of the patch size.

\textbf{Randomization:}  In our experiments, we observe that without giving access to the patch dictionaries (\ie the grey-box setting), the D3 algorithm is successful in defending against the adversarial attack.
However, the defense is weaker when the adversary has access to the dictionary.
This observation motivates us to add randomization to our transformation function, $T(.)$, when the adversary has full access to the D3 algorithm (white-box setting).
By adding randomization, even though the adversary can access the dictionary, exact atoms used for reconstruction will not be available.
We add following efficient randomization schemes (both training and test time) to our defense:
\vspace{-.1in}
\begin{itemize}
  \setlength\itemsep{0.01em}
  \item We randomize over the columns of the dictionaries by randomly selecting one fifth of the atoms in the patch dictionaries while denoising a patch.
  \item We further randomize by first selecting the top-$2$ most correlated atoms from the patch dictionary and randomly picking one of those two.
\end{itemize}

\section{Experiments}
\label{sec:experiments}
We conduct experiments on the $1000$-class
ImageNet dataset~\cite{imagenet_cvpr09} with a ResNet-50~\cite{KaimingHeResnet}.
We evaluate defense against the following diverse set of attack algorithms --
\textbf{FGSM}~\cite{goodfellow15} which is a one-step attack where gradient direction is multiplied by the norm of the noise,
\textbf{DeepFool}~\cite{moosavi16} that is an iterative attack ensuring that the network output is changed,
\textbf{CWattack}~\cite{carlini17} that solves an optimization problem to find the adversarial perturbation, and
\textbf{UAP}~\cite{moosavi17} as a transferable attack that computes a universal noise for all the images.
We compute the adversarial perturbation in floating point and scale it to have a fixed $\ell_2$ norm, \ie $\frac{\|\boldsymbol v\|_2}{\|\boldsymbol x\|_2} = 0.06$.
This noise is added to the image after it has been converted to floating point and normalized.
This experimental setup is described in~\cite{guo17}.
We compare the D3 algorithm with several image transformation based defenses proposed in ~\cite{guo17}
and show significant improvement in classification accuracy.
Furthermore, we analyze the tradeoff between the clean image accuracy and the robustness against the attacks.
We study the effect of the hyper-parameter values and show how they can be tuned to improve the accuracy or the robustness.
We use a pre-trained network on the original image space $\boldsymbol x$ and fine-tune it on the transformed images $T(\boldsymbol x)$.
We set $\epsilon=0.85$ for dictionary reconstruction and discuss the effects of other hyper-parameters in Section~\ref{sec:hyp_para_sel}.

\paragraph{Tradeoff between Accuracy and Robustness:}
Hyper-parameters of the D3 algorithm can be used to control the quality of image reconstruction.
For example, as we increase the sparsity, $\kappa$, more details are preserved and the accuracy on the reconstructed clean images, $T(\boldsymbol x)$, is improved.
However, if the image has been corrupted with adversarial noise, increasing $\kappa$ reconstructs more noise,
and the accuracy on the reconstructed noisy images, $T(\boldsymbol x + \boldsymbol v)$,  eventually decreases. 

\begin{figure*}[t]
  \centering
  \subfigure[]{
  \includegraphics[width=0.8\columnwidth]{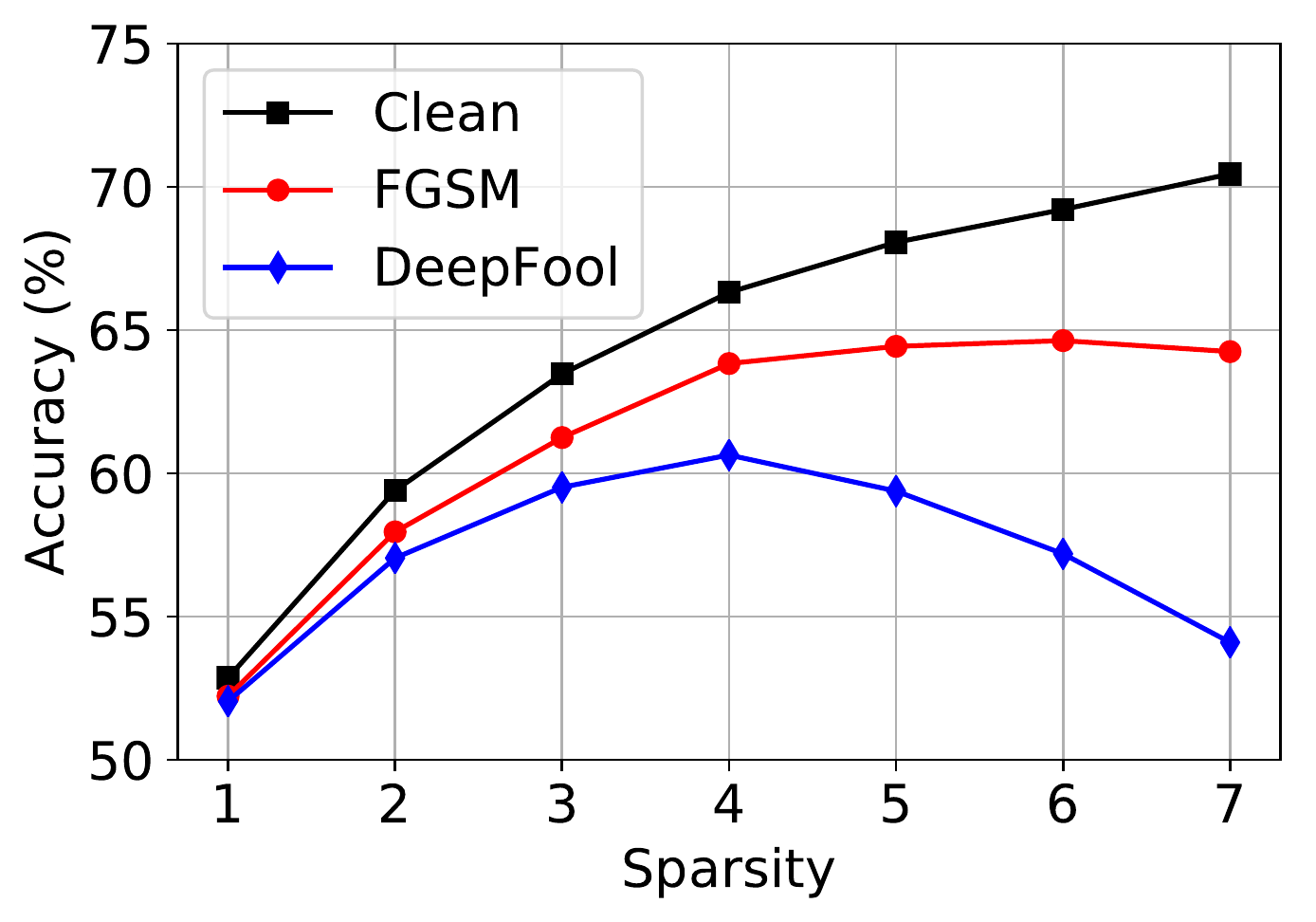}
  }
  \hskip0.4in
  \subfigure[]{
  \includegraphics[width=0.928\columnwidth]{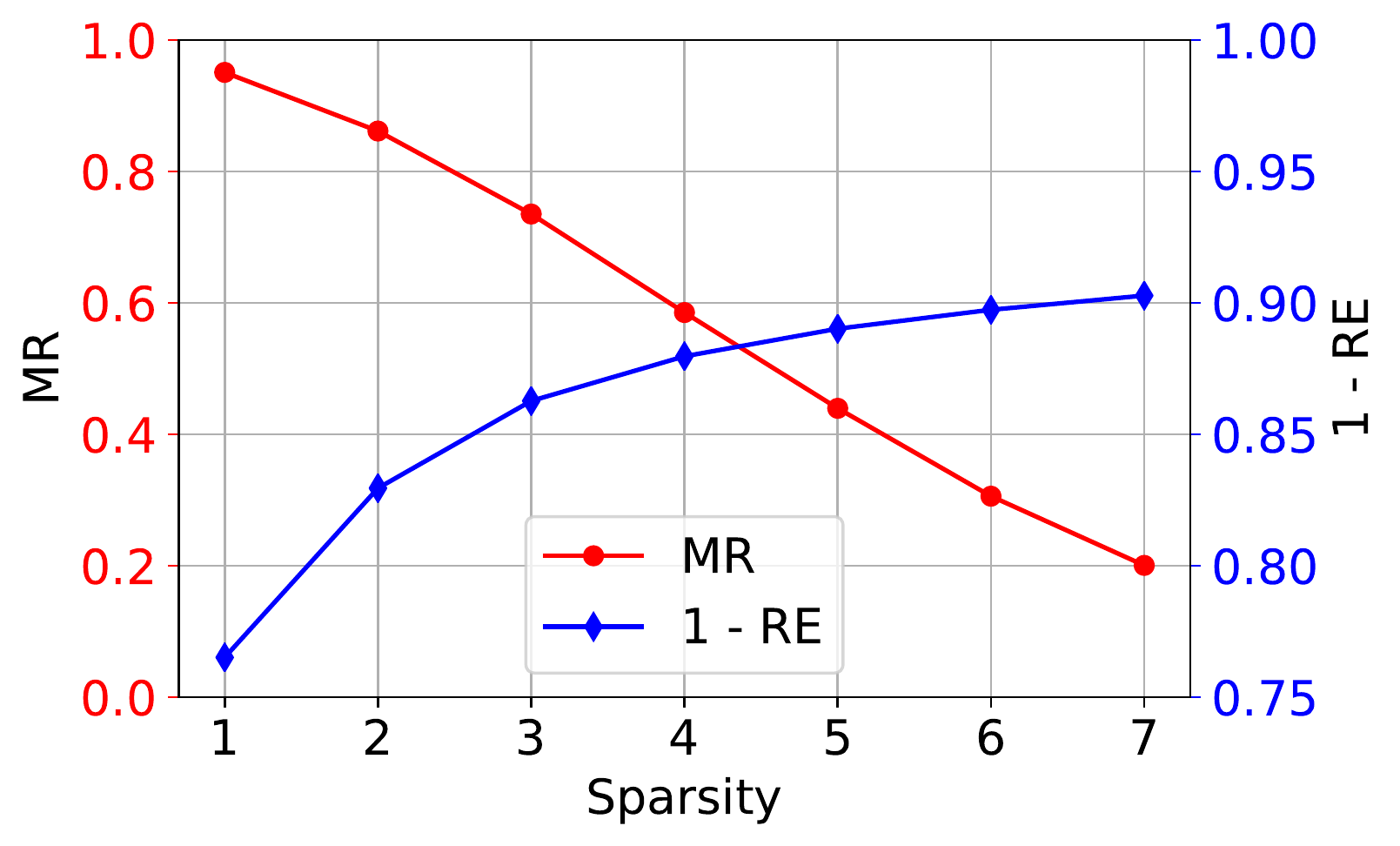}
  }
  \vskip -0.1in
  \caption{  (a) Tradeoff between the clean image accuracy and the robustness of the classifier with increasing sparsity.
  (b) Matchine rate (MR) and reconstruction quality ($1$ - RE) for the same sparsity values.
  From these two figures, we can see that clean image accuracy and the reconstruction quality increase with sparsity.
  The MR decreases with sparsity, causing the classifier to be less robust.
  The classifier performance on adversarially perturbed images is low with small sparsity due to low reconstruction quality, and is also low with large sparsity because of low matching-rate.
  The optimal performance is achieved at an intermediate value (\eg $\kappa=4$ for DeepFool attack).
  Due to computational constraints, we reduce the patch overlap to $8$ pixels for this analysis which improves the reconstruction speed by 4x.
  }
  \label{fig:sparsity}
\end{figure*}

We plot the accuracy on clean and noisy images with a grey-box attack in Figure~\ref{fig:sparsity}(a).
As we can see from the plots, the classification accuracy on the clean images consistently improves with sparsity.
The accuracy on the noisy images is low with small sparsity and increases in the beginning as the reconstruction quality improves.
After increasing the sparsity to a large value, the noise also starts getting reconstructed, and the classification accuracy drops.
Hence, there is a ``sweet spot'' (for example, $\kappa=4$ for DeepFool) for optimal defense against the attack, and sparsity can be used as a tradeoff parameter.

Figure~\ref{fig:sparsity}(b) shows the matching-rate (MR) and the reconstruction-quality ($1$ - RE) for the same sparsity values.
We see that the reconstruction quality improves with sparsity, and is correlated with the classification accuracy on the clean images.
However, the matching-rate decreases as sparsity is increased, causing the network to be less robust to adversarial noise.
In our experiments, we find that dictionary size also plays a role in the accuracy and robustness tradeoff.
A larger dictionary improves the accuracy on the clean images because the images are better reconstructed.
A smaller dictionary generally improves the robustness because the dictionary atoms are, on average, farther apart.
This correlation between the (MR, 1-RE) and the classifier's performance enables us to efficiently study the effects of hyper-parameters.

Based on this analysis, we pick three settings to evaluate our algorithm.
In the first setting, we set high sparsity and choose a large dictionary ($\kappa = 5$, dictionary size $\eta = 40k$).
This setting encourages high accuracy on the clean images, but is less robust against adversarial attacks.
In the second setting, we reduce the dictionary size while keeping the sparsity the same ($\kappa = 5$, dictionary size $\eta = 10k$).
In the third setting, we set the dictionary size $\eta = 10k$, and reduce the sparsity to $4$, further improving the robustness.
Next, we describe three types of attacks (black-box, grey-box, and white-box) and compare our results with the state-of-the-art.

\paragraph{Black-box Attack:}
In this attack, the adversary does not have access to the network parameters, or the defense mechanism.
We use a separate ResNet-50 network and compute the adversarial perturbations using original images, $\boldsymbol x$.
As mentioned earlier, the $\ell_2$-norm of the noise is set to $0.06$ and it is added to the floating point image.
As shown in Table~\ref{tb:black-box-results}, all the methods are robust to these attacks as this attack is the easiest to defend.
Note that here we assume that the adversary knows about the network architecture but not the exact parameters.
We also compare with the best method~\cite{liao2017defense} in the \emph{NIPS2017 challenge}.
This method used a different base network (InceptionV3) which had a higher clean image (no-attack) accuracy of $76.6\%$.
We used their pre-trained network/defense algorithm and evaluated it with our experimental setup.
Compared to D3, it had a significantly lower accuracy under DeepFool attack ($59.8\%$) and slightly better accuracy ($71.9\%$) under FGSM attack.
Note that better performance on FGSM attack may be due to their defense being trained using adversarially perturbed images with FGSM.

\begin{table*}[t]
\caption{Top-1 classification accuracy (\%) with black-box attacks ($1000$ ImageNet classes).}
\label{tb:black-box-results}
\vspace{-0.15in}
\begin{center}
\begin{small}
\begin{sc}
\begin{tabular}{lcccccr}
\toprule
Method & No Attack & DeepFool & CWAttack & FGSM & UAP \\
\midrule
D3 ($\eta=40k, \kappa=5$) & ${71.8}$ & $63.1$ & $63.3$ & $\boldsymbol{68.6}$ & $\boldsymbol{71.5} $      \\
D3 ($\eta=10k, \kappa=5$)  &  $70.8$& $64.6$ & $64.8$ & $68.3$ & $70.3$       \\
D3 ($\eta=10k, \kappa=4$)  &  $69.0$& ${64.8}$  & $65.0$ & $67.1$ & $68.9$       \\

Quilting~\cite{guo17} & $70.1$ &  $65.2$ & $64.1$ & $65.5$ & -        \\
TVM + Quilting~\cite{guo17} & $\boldsymbol{72.4}$&  $65.8$  & $64.0$ & $65.7$ & -       \\
Cropping + TVM + Quilting~\cite{guo17} & $72.1$ &  $\boldsymbol{67.1}$   & $\boldsymbol{65.3}$ & $66.7$ & -       \\
\bottomrule
\end{tabular}
\end{sc}
\end{small}
\end{center}
\vspace{-0.1in}
\end{table*}

\begin{table*}[t]
\caption{Top-1 classification accuracy (\%) with grey-box attacks ($1000$ ImageNet classes).}
\label{tb:grey-box-results}
\vspace{-0.15in}
\begin{center}
\begin{small}
\begin{sc}
\begin{tabular}{lcccccr}
\toprule
Networks & No Attack & DeepFool & CWAttack & FGSM & UAP \\
\midrule
D3 ($\eta=40k, \kappa=5$) &$71.8$ & $58.7$ & $57.9$ & $67.3$ & $\boldsymbol{71.5}$        \\
D3 ($\eta=10k, \kappa=5$)  &  $70.8$ & $62.3$ & $62.4$ & ${67.5}$ & $70.4$         \\
D3 ($\eta=10k, \kappa=4$)  &  $69.0$ & $\boldsymbol{64.4}$ & $\boldsymbol{64.5}$ & $67.1$ & $68.7$         \\
Quilting~\cite{guo17}  &  $66.9$ & $34.5$ & $30.5$ & $39.6$ &-        \\
Cropping~\cite{guo17}  &  $65.4$ & $44.9$ & $41.1$ & $49.5$ &-        \\
TVM~\cite{guo17}  &  $66.3$ & $44.7$ & $48.4$  & $31.4$ &-        \\
 Ensemble Training~\cite{tramer17,guo17}\footnotemark[1]
          &  $\boldsymbol{80.3}$ & $1.8$ & $22.2$ & $\boldsymbol{69.2}$ & -         \\

\bottomrule
\end{tabular}
\end{sc}
\end{small}
\end{center}
\vspace{-0.1in}
\end{table*}

\begin{table*}[!h]
\caption{Top-1 classification accuracy (\%) with white-box attacks ($1000$ ImageNet classes). }
\label{tb:white-box-results}
\vspace{-0.15in}
\begin{center}
\begin{small}
\begin{sc}
\begin{tabular}{lcccc}
\toprule
Networks & No Attack & DeepFool & CWAttack  & FGSM\\ 
\midrule

Xie et al.~\cite{XieIclr2018, athalye18a}\footnotemark[2] & - & - & $0.0$  & - \\
Quilting~\cite{guo17, athalye18a}\footnotemark[2] & - & - & $0.0$  & - \\
D3 ($\eta=40k, \kappa=5$) & $\boldsymbol{70.8}$ & $27.5$ & $27.1$ & ${54.5}$     \\
D3 ($\eta=10k, \kappa=5$)   & $69.8$&  ${31.0}$   & $30.9$ & ${55.8}$         \\
D3 ($\eta=10k, \kappa=4$)  & $68.2$ &  $\boldsymbol{34.3}$ & $\boldsymbol{34.4}$  & $\boldsymbol{56.9}$        \\

\bottomrule
\end{tabular}
\end{sc}
\end{small}
\end{center}
\end{table*}

\paragraph{Grey-box Attack:}
In this setting, the adversary does not have access to the defense mechanism but knows about the network weights.
We compute all the noise patterns using gradients of the fine-tuned network evaluated at the original images, $\boldsymbol x$.
This setting is the same as the ``white-box'' setting in~\cite{guo17}, so we compare our results in grey-box setting to their white-box setting.
Without any defense mechanism, the attacks are very successful in this setting: FGSM reduces the classification accuracy to $6.2\%$, DeepFool reduces it to $9.2\%$, and UAP reduces it to $20.8\%$.
Our results in Table~\ref{tb:grey-box-results} show that we perform significantly better than the state-of-the-art.
We also note that as we decease the dictionary size $\eta$ or the sparsity $\kappa$, the robustness of the classifier improves.
For example, with $\eta=40k, \kappa = 5$, the DeepFool accuracy is $58.7\%$ which improves to $64.4\%$ by reducing $\eta$ to $10k$ and sparsity to $4$.
As before, we do not use any adversarial examples while fine-tuning the network.
Ensemble training~\cite{tramer17}\footnotemark[1] does better on FGSM attack ($69.2\%$ accuracy) by training the network with FGSM adversarial examples.
But the trained model performs poorly with DeepFool attack, resulting only in $1.8\%$ accuracy.
\footnotetext[1]{The results of the ensemble training algorithm are taken from~\cite{guo17}.}

\paragraph{White-box Attack:}
Since the D3 transformation function is not differentiable, we cannot fool the network using gradient-based attacks (FGSM, DeepFool, CWAttack, or UAP).
We compute the adversarial noise, $\boldsymbol v$, using the fine-tuned network weights while the gradients are computed at the transformed image, $T(\boldsymbol x)$.
This noise, $\boldsymbol v$, is added to the image $\boldsymbol x$.
This attack is same as  BDPA of~\cite{athalye18a}.
Under this challenging setting, the DeepFool attack reduces the classification accuracy to $13.0\%$, and the FGSM attack reduces it to $34.4\%$.
We can make D3 more robust by adding randomization, which prevents the attacker from accessing the exact atoms used for reconstruction, as described in Section~\ref{sec:D3_denoising_algo}.
Classification accuracies using D3 with randomization are shown in Table~\ref{tb:white-box-results} which shows significant improvement over the deterministic version.
As with the other attack types, the algorithm is more robust with a smaller dictionary and lower sparsity.
We achieve accuracy of $13\%$ without randomization and $34.4\%$ with randomization on the BPDA compared to the $0\%$ accuracy reported in ~\cite{athalye18a}.
We hypothesize that this improvement is partially due to the fact that our transformation radically reduces the effective dimensionality of the input images.
We are able to use a bigger patch size $16\times16$ (even upto $32 \times 32$) which effectively limits the search space for an adversary.

\footnotetext[2]{The results of the BPDA attack are taken from~\cite{athalye18a}.}

\subsection{Hyper-parameters selection}
\label{sec:hyp_para_sel}
The proposed D3 defense has the following hyper-parameters: patch-size, $P$, sparsity, $\kappa$, dictionary size, $\eta$,
and minimum distance between two dictionary atoms, $\epsilon$.
They affect the overall performance in different ways.
For example, as we increase the patch size, the robustness of the classifier improves while it becomes less accurate.
Increasing the minimum angle between two atoms, i.e. $\epsilon$, also improves the robustness but degrades the classifier accuracy.
However, we recommend keeping $\epsilon$ high because it improves the robustness more than it hurts accuracy on the clean images.

\begin{figure}
\begin{minipage}[t]{.48\textwidth}
\centering
\includegraphics[width=1\textwidth]{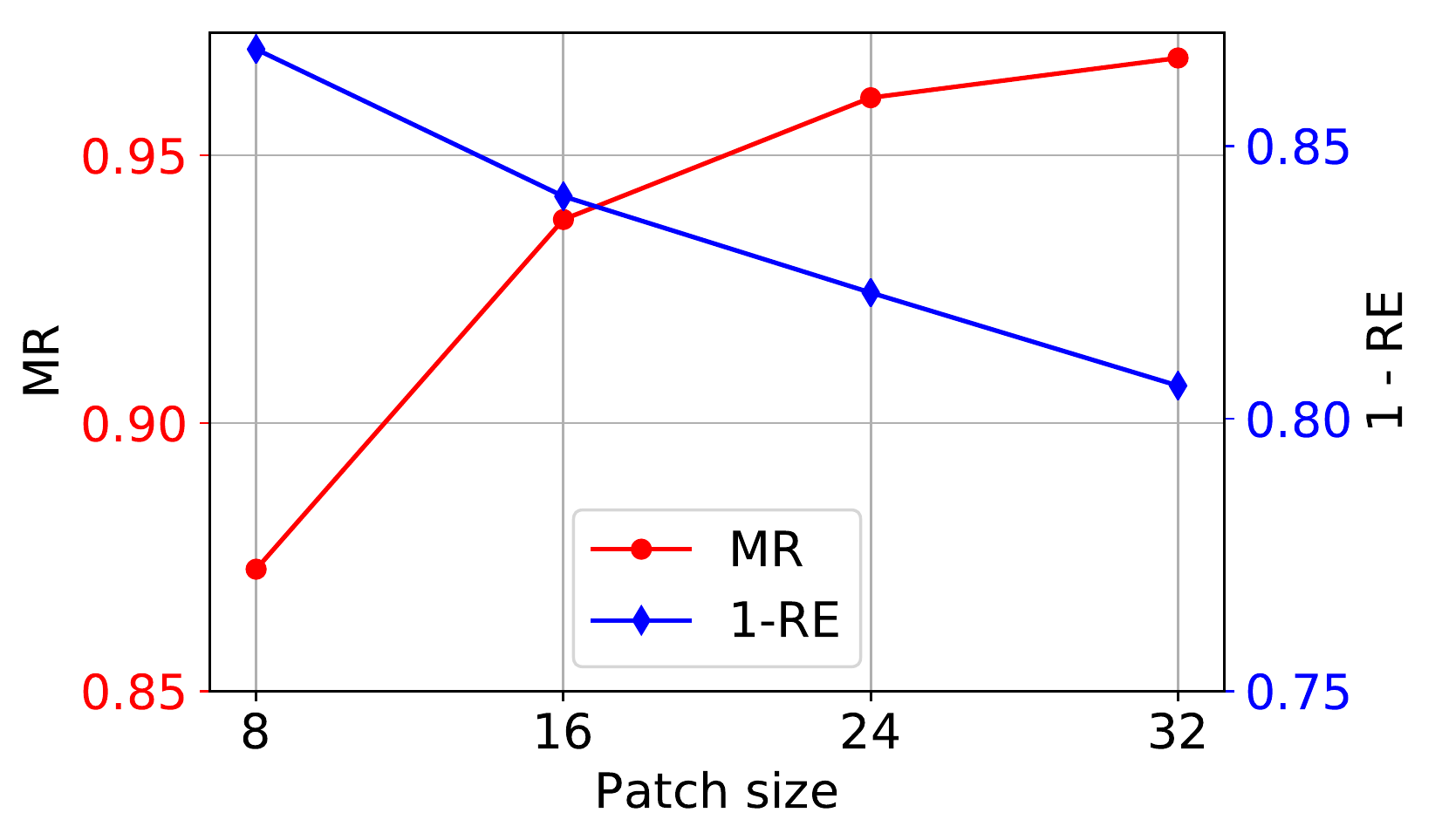}
\caption{Effect of patch size ($P$) on matching-rate and reconstruction-quality ($1$ - RE).
Increasing $P$ improves the matching-rate (robustness) but reduces the reconstruction-quality.
}
\label{fig:patchsize}
\end{minipage}
\hfill
\begin{minipage}[t]{.48\textwidth}
\centering
\includegraphics[width=1\textwidth]{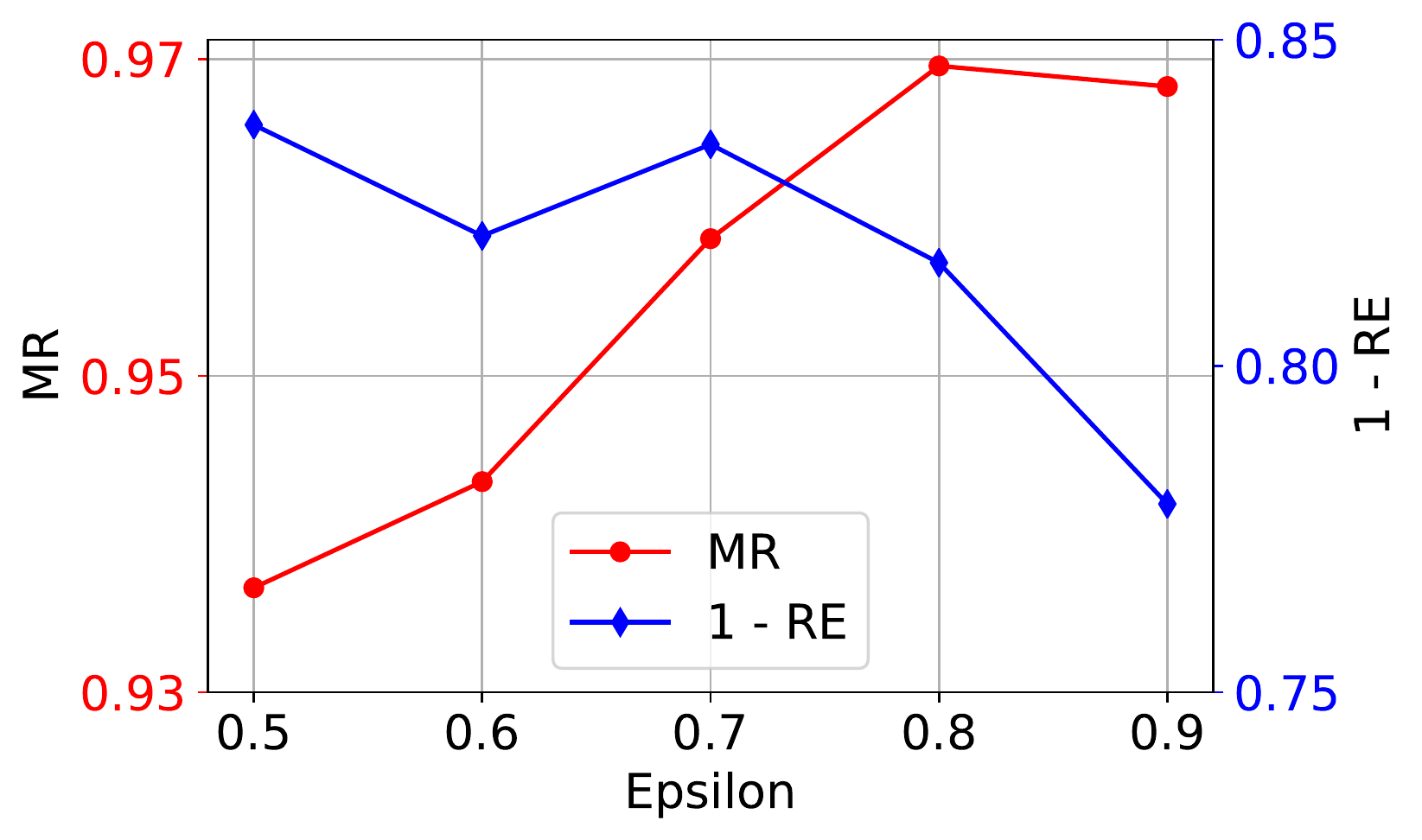}
\caption{
Matching-rate improves as $\epsilon$ increases or correlation among dictionary atoms decreases.
The reconstruction-quality ($1 - \text{RE}$) decreases with increasing $\epsilon$.
}
\vspace{-0.2in}
\label{fig:epsilon}
\end{minipage}
\end{figure}

We analyze the effect of $\kappa$ on accuracy in Figure~\ref{fig:sparsity},
and show how the accuracy and the robustness are correlated with the matching-rate (MR) and the reconstruction-quality ($1$ - RE).
We show the effect of different patch sizes ($P=8,16,24,32$) in Figure~\ref{fig:patchsize} on matching-rate and reconstruction-quality.
As we increase the patch size, the matching-rate increases that is the robustness of the classifier improves.
However, the reconstruction-quality ($1 - \text{RE}$) decreases making the classifier less accurate.

\begin{figure}[t]
\centering

\subfigure[$ \kappa=1$]{
\includegraphics[width=.18\textwidth]{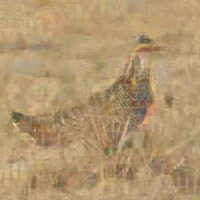}
}
\subfigure[$\kappa=3$]{
\includegraphics[width=.18\textwidth]{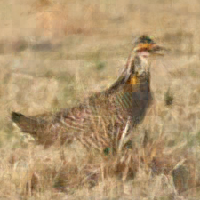}
}\\
\subfigure[$\kappa=5$]{
\includegraphics[width=.18\textwidth]{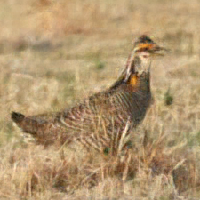}
}
\subfigure[Original Image]{
\includegraphics[width=.18\textwidth]{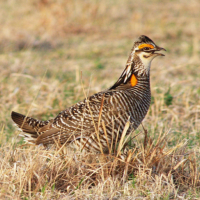}
}

\vspace{-0.1in}
\caption{Denoised images with varying sparsity.}
\vspace{-0.2in}
\label{fig:example_imgs_o_k}
\end{figure}

Next, we study the effect of $\epsilon$ in Figure~\ref{fig:epsilon}.
As expected, increasing the minimum angle between two atoms, increases the matching-rate and will improve the robustness.
We also find that the reconstruction-quality ($1 - \text{RE}$) decreases, thus decreasing the classifier accuracy.
Figure~\ref{fig:example_imgs_o_k} shows example images with different sparsity values.

\captionsetup[subfigure]{labelformat=empty}

\paragraph{Effect of Task-Complexity:}
For simpler tasks, the D3 algorithm can reduce more information from the images while maintaining the classification accuracy and the robustness to attack.
To evaluate our defense mechanism on a lower complexity task, we train a classifier on $50$ randomly selected ImageNet classes.
Under black-box attack, in the worst case the performance reduces from $93.2\%$ to $91.8\%$ and under grey-box attack it reduces to $89.7\%$.
In the white-box setting, the top-1 classification accuracy on adversarial images drops to $70.9\%$ from $91.7\%$ (on clean images).
We also evaluated our algorithm on CIFAR-10 dataset, another lower complexity task, and compared with many state-of-the-art methods given in~\cite{song2018pixeldefend} under FGSM attack.
Our algorithm achieved $87\%$ clean accuracy and $80\%$ accuracy under the attack.
Our attack-agnostic defense is better than all the defense methods but the one that specialized specifically for FGSM attack by adding adversarial images to the training.
The complete evaluation on small ImageNet and CIFAR-10 datasets are included in Appendices A and B, respectively.



\section{Conclusion}
We described a novel patch-based denoising algorithm to improve the robustness of classifiers, trained on large-scale datasets, against adversarial perturbations.
We developed two proxy metrics (MR and RE) to help guide us in designing the algorithm.
The design of our denoising (D3-MP), and patch selection (D3-DL) algorithms encourages high matching-rate between the clean patch and the corresponding noisy patch.
We provided a thorough study of the tradeoff between clean image accuracy and robustness against the attacks.
We proposed an efficient randomization schemes to improve the robustness against white-box attacks.
Our evaluation on the ImageNet show that our defense mechanism provides state-of-the-art results.

\small
\bibliographystyle{plain}
\bibliography{refs}
\raggedbottom 
\pagebreak

\subsection*{Appendix A: Experiments on $50$-class ImageNet}
For simpler tasks, the D3 algorithm can reduce more information from the images while maintaining the classification accuracy and the robustness to attack.
To evaluate our defense mechanism on a lower complexity task, we train a classifier on $50$ randomly selected ImageNet classes.
We find that the proposed defense (D3) is successful on this lower complexity task.

As shown in Tables~\ref{tb:black-box-results-50-classes} and \ref{tb:grey-box-results-50-classes}, black and grey-box attacks lower the top-1 classification accuracy by less than $4\%$.
In the most challenging white-box setting including the DeepFool attack, the top-1 classification accuracy drops to $70.9\%$ from $91.7\%$ on clean images.
This $20.8\%$ drop compares favorably with the $35.3\%$ loss in accuracy for the $1000$-class classifier under the same setting.

\begin{table}[h]
\caption{Top-1 classification accuracy (\%) on black-box attacks ($50$ ImageNet classes)}
\label{tb:black-box-results-50-classes}
\vspace{-0.05in}
\begin{center}
\begin{small}
\begin{sc}
\begin{tabular}{lcccr}
\toprule
\scriptsize{Networks} & \scriptsize{No Attack} & \scriptsize{DeepFool} & \scriptsize{FGSM} \\
\midrule
D3 ($\eta=40K, \kappa=5$) &$\boldsymbol{94.0}\%$ & $\boldsymbol{92.1}$ & $\boldsymbol{93.3}$      \\
D3 ($\eta=10K, \kappa=5$)  &  $93.9$ & $91.8$ & $93.1$       \\
D3 ($\eta=10K, \kappa=4$)  &  $93.2$ & $91.8$ & $92.2$     \\
\bottomrule
\end{tabular}
\end{sc}
\end{small}
\end{center}
\end{table}

\begin{table}[h]
\caption{Top-1 classification accuracy (\%) on grey-box attacks ($50$ ImageNet classes)}
\label{tb:grey-box-results-50-classes}
\vspace{-0.05in}
\begin{center}
\begin{small}
\begin{sc}
\begin{tabular}{lcccr}
\toprule
\scriptsize{Networks} & \scriptsize{No Attack} & \scriptsize{DeepFool} & \scriptsize{FGSM} \\
\midrule
D3 ($\eta=40K, \kappa=5$) &$\boldsymbol{94.0}$ & $86.0$ & $91.8$      \\
D3 ($\eta=10K, \kappa=5$)  &  $93.9$ & $88.0$ & $\boldsymbol{91.8}$       \\
D3 ($\eta=10K, \kappa=4$)  &  $93.2$ & $\boldsymbol{89.7}$ & $91.7$     \\
\bottomrule
\end{tabular}
\end{sc}
\end{small}
\end{center}
\end{table}

\begin{table}[h]
\caption{Top-1 classification accuracy (\%) on white-box attacks ($50$ ImageNet classes)}
\label{tb:white-box-results-50-classes}
\vspace{-0.05in}
\begin{center}
\begin{small}
\begin{sc}
\begin{tabular}{lcccr}
\toprule
\scriptsize{Networks} & \scriptsize{No Attack} & \scriptsize{DeepFool} & \scriptsize{FGSM} \\
\midrule
D3 ($\eta=40K, \kappa=5$) &$\boldsymbol{93.3}$ & $66.2$ & $85.4$      \\ 
D3 ($\eta=10K, \kappa=5$)  &  $92.8$ & $66.4$ & ${85.4}$       \\ 
D3 ($\eta=10K, \kappa=4$)  &  $91.7$ & $\boldsymbol{70.9}$ & $\boldsymbol{85.8}$     \\ 
\bottomrule
\end{tabular}
\end{sc}
\end{small}
\end{center}
\end{table}

\subsection*{Appendix B: Experiments on CIFAR-10 dataset}
Table~\ref{tb:cifar-10} compares the D3 algorithm withe several recent defense algorithms under FGSM attack.
Evaluation of the defense algorithms is taken from PixelDefend~\cite{song2018pixeldefend}(PD) paper.
The D3 performs better than all the methods except Adversarial FGSM which is specifically trained using FGSM perturbed images, while our method is attack-agnostic.

\begin{table}[h]
\caption{Classification accuracy (\%) CIFAR-10 dataset.}
\label{tb:cifar-10}
\vspace{-0.05in}
\begin{center}
\begin{footnotesize}
\begin{sc}
\begin{tabular}{llcr}
\toprule
{Network} & Training/Defense & {No Attack}  & {FGSM} \\
\midrule
ResNet & Normal   & $92$ & $11$      \\ 
VGG & Normal   & $89$ & $30$      \\ 
ResNet & Adversarial FGSM   & $91$ & $91$      \\ 
ResNet & Adversarial BIM   & $87$ & $34$      \\ 
ResNet & Label Smoothing   & $92$ & $28$      \\ 
ResNet & Feature Squeezing   & $84$ & $18$      \\ 
ResNet & Adversarial FGSM   & $86$ & $55$      \\ 
        & \;\;+ Feature Squeezing & & \\
ResNet & Normal  & $85$ & $24$      \\ 
       & \;\;+ PixelDefend & & \\
VGG & Normal + PixelDefend   & $82$ & $52$      \\ 
ResNet & Adversarial FGSM   & $88$ & $67$      \\ 
      & \;\;+ PixelDefend & & \\
ResNet & Adversarial FGSM    & $90$ & $67$      \\ 
      & \;\;+ Adaptive PixelDefend & & \\
ResNet & D3   & $87$ & $80$      \\ 
\bottomrule
\end{tabular}
\end{sc}
\end{footnotesize}
\end{center}
\end{table}

\raggedbottom 

\end{document}